\documentclass[conference]{llncs}
\usepackage{cite}
\usepackage{amsmath,amssymb,amsfonts}
\usepackage{graphicx}
\usepackage{textcomp}
\usepackage{xcolor}
\usepackage[T1]{fontenc}
\usepackage[latin9]{inputenc}
\usepackage{wrapfig}
\makeatletter

\usepackage{algorithm,algpseudocode}
\usepackage[inline]{enumitem}

\@ifundefined{showcaptionsetup}{}{%
 \PassOptionsToPackage{caption=false}{subfig}}
\usepackage{subfig}
\makeatother

\def\BibTeX{{\rm B\kern-.05em{\sc i\kern-.025em b}\kern-.08em
    T\kern-.1667em\lower.7ex\hbox{E}\kern-.125emX}}
\begin{document}

\title{Intuitive Physics Guided Exploration for Sample Efficient Sim2real Transfer\thanks{This research was partially funded by the Australian Government through the Australian Research Council (ARC). Prof Venkatesh is the recipient of an ARC Australian Laureate Fellowship (FL170100006).}
}

\author{Buddhika Laknath Semage\textsuperscript{*} \and Thommen George Karimpanal \and Santu Rana \and Svetha Venkatesh}
\institute{Applied Artificial Intelligence Institute, Deakin University, Geelong, Australia
\email{bsemage@deakin.edu.au\textsuperscript{*}}}
\maketitle

\begin{abstract}
Physics-based reinforcement learning tasks can benefit from simplified physics simulators as they potentially allow near-optimal policies to be learned in simulation. However, such simulators require the latent factors (e.g. mass, friction coefficient etc.,) of the associated objects and other environment-specific factors (e.g. wind speed, air density etc.,) to be accurately specified, without which, it could take considerable additional learning effort to adapt the learned simulation policy to the real environment. 
As such a complete specification can be impractical, in this paper, we instead, focus on learning task-specific estimates of latent factors which allow the approximation of real world trajectories in an ideal simulation environment. 
 Specifically, we propose two new concepts: a) action grouping - the idea that certain types of actions are closely associated with the estimation of certain latent factors, and; b) partial grounding - the idea that simulation of task-specific dynamics may not need precise estimation of all the latent factors. 
We first introduce intuitive action groupings based on human physics knowledge and experience, which is then used to design novel strategies for interacting with the real environment. Next, we describe how prior knowledge of a task in a given environment can be used to extract the relative importance of different latent factors, and how this can be used to inform partial grounding, which enables efficient learning of the task in any arbitrary environment. We demonstrate our approach in a range of physics-based tasks, and show that it achieves superior performance relative to other baselines, using only a limited number of real-world interactions.

\end{abstract}

\section{Introduction}




One of the defining characteristics of human learning is the ability
to leverage domain knowledge to learn quickly and with very little data. This is a key characteristic many
machine learning models, including reinforcement learning (RL), lack\cite{loula2018rearranging,DBLP:conf/icml/LakeB18}.
For example, to attain similar levels of performance, deep Q-networks (DQN) \cite{mnih2015human} have been reported to require $50$ million frames of experience
($38$ days of gameplay in real-time) , which stands in stark contrast to the less than $15$ minutes of gameplay required by human players \cite{Tsividis2017-by}. This difference has largely been attributed to the fact that humans leverage prior knowledge \cite{dubey2018investigating} for such tasks, allowing for more efficient exploration during learning.

\begin{wrapfigure}{r}{0.5\textwidth}
  \begin{center}
    \includegraphics[width=0.4\textwidth]{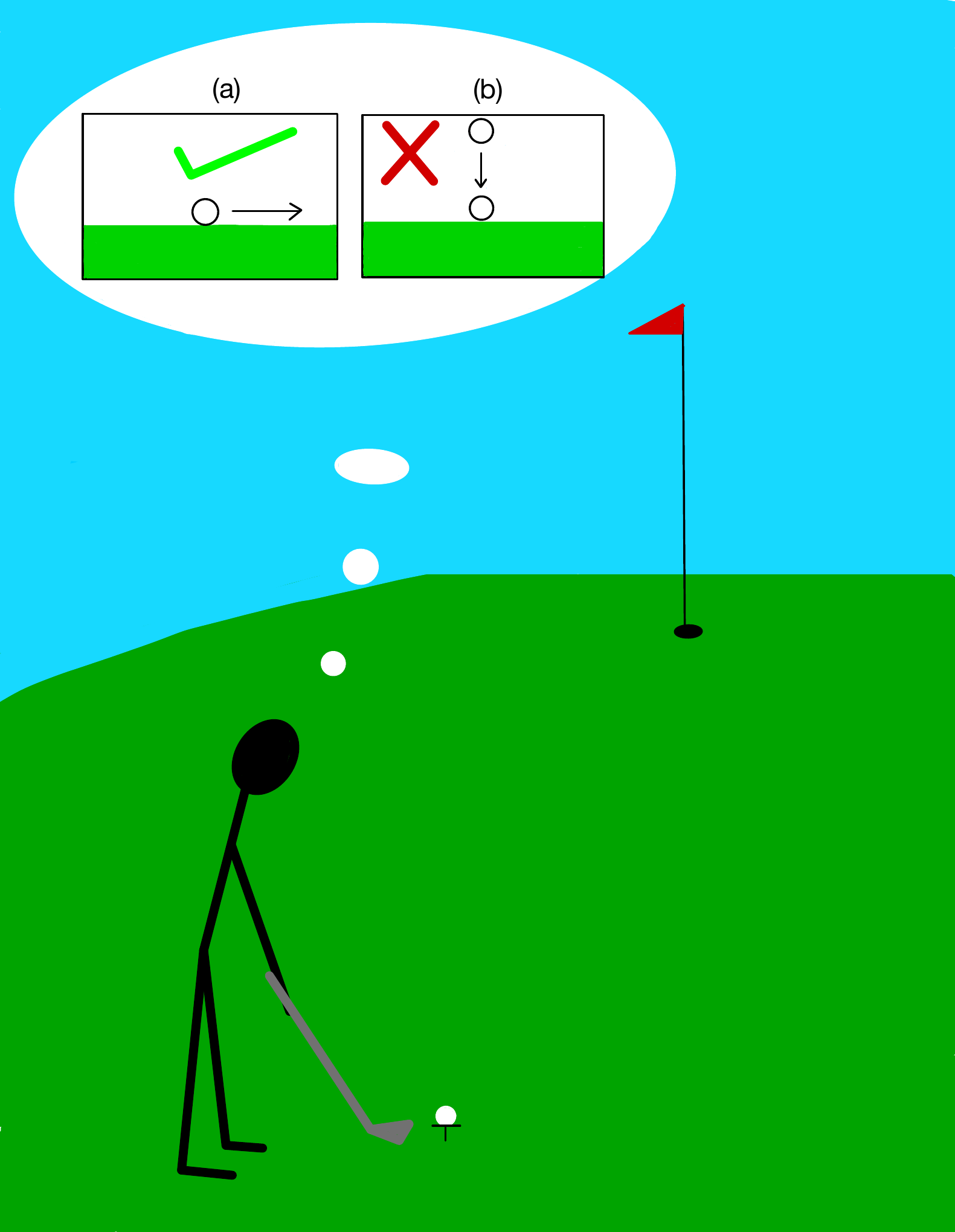}
  \end{center}
  \caption{Action groupings represented by (a) rolling and (b)collision action types. Due to partial grounding, only rolling needs to be explored for a `putt' shot in golf.}
\label{fig:teaser}
\end{wrapfigure}
 
In physics-based problems,
one can leverage the known laws governing the physical world, as described by classical mechanics. This structured prior knowledge can be assumed to be encapsulated in physics simulation softwares, and can be leveraged for training artificial agents on physics based tasks \cite{Wu2015-sq,farchy2013humanoid}. For example, when training a robot to play golf, with good estimates of the latent factors (eg., mass, friction) associated with the ball,
club and turf, a physics simulator would be able to compute the approximate ball trajectory. Such approximations allow the possibility of learning approximate policies solely in the simulator, obviating the need for extensive interactions with the real world. 

Recently proposed methods such as domain randomisation (DR) \cite{8202133,DBLP:conf/rss/SadeghiL17} operate precisely on such a basis, where a simulation policy is first trained on a variety of simulation environments, each of which is generated using randomly sampled latent factors. The idea is that the policy obtained as a result of training on such a diverse set of simulation environments would easily transfer to the real environment, as the latter can be expected to be close to one or more of the environments experienced in simulation. 

While DR produces robust policies against environmental variations, the resulting policy does not behave optimally in the given real environment. To achieve this, DR agents require further training using real-world interactions, which can be prohibitively expensive. Instead, using fewer interactions, if we were able to learn latent factors to closely reproduce realistic task-specific trajectories in simulation, the corresponding simulation policy would transfer seamlessly to the real environment. 
Here, we propose an \emph{Intuitive Physics Learning Workflow
(IPLW)} to learn such a simulation policy (a) by using intuitive \emph{action groupings}, according to which, specific latent factors are associated with specific types of actions and (b) by using the fact that depending on the task, certain latent factors may be more useful to estimate than others, an idea we term \emph{partial grounding}.   
For example, through intuitive action groupings, we may assume that rolling and collision actions are typically associated with estimating friction and coefficient of restitution (CoR) respectively. Partial grounding is elucidated by the fact that depending on the task under consideration, it may not be necessary to treat each latent factor with equal importance. That is, for a rolling-intensive task (e.g. a `putt' shot in golf, as shown in Fig. \ref{fig:teaser}), one need not spend considerable effort on estimating CoR, as it is likely to be less relevant to the task. Similar ideas proposed in concepts such as observational dropout \cite{freeman2019learning} have previously been shown to be useful for model estimation. 


To implement IPLW, actions are first ranked according to simple heuristics such as the number of collisions or the rolling time experienced during a trajectory. The top-ranked subsets of these actions are then classified as either collision or rolling action types, each of which is associated with the estimation of a specific latent factor (i.e., CoR and friction respectively). In this way, we use knowledge of physics relationships to remove less informative actions from consideration, which constrains the action space significantly. Furthermore, such action categorization fits naturally with the idea of partial grounding, as it allows agents to interact with the environment based on the specific latent factors to be estimated. In implementing partial grounding, we use task-specific knowledge from a given environment to accelerate learning in another environment, which may differ in terms of the unmodeled factors (e.g. air resistance, surface roughness etc.). 
Specifically, we first characterize the known environment through its empirically estimated transfer performances for various latent factor values on an ideal simulator (i.e. without the unmodeled factors). Using this performance surface, we compute the gradients of the transfer performance with respect to latent factors, and use these as exploration cues to choose specific action types when exploring the new environment. 

In summary, the main contributions of this work are:
\begin{itemize}
\item Proposing a paradigm of intuitively biased action selection for task-specific grounding in physics based RL tasks. 
\item Developing a novel interaction strategy that estimates relevant latent factors to closely approximate real world trajectories in the simulator.
\item Empirical demonstrations of the framework through superior jump start improvements in a range of different physics based environments.
\end{itemize}


\section{Related Work}
Studying human understanding of physics dynamics and the learning
process has provided useful insights that have been used to improve
physics related machine learning models. These models can be classified
into two categories as top-down and bottom-up designs: bottom-up designs
attempt to learn physics representations without providing external
biases \cite{Fragkiadaki2016-qf,Grzeszczuk1998-na,Janner2018-de},
whereas top-down models try to build improved physics models using
the physics knowledge we possess. Our study adopts a top-down design
aiming to construct a better bias.

\emph{Galileo} \cite{Wu2015-sq} was an early attempt at learning
simple physics dynamics such as sliding of solid blocks on a ramp.
It formulated a model that comprised a physics software engine and
utilised a Monte Carlo sampling approach to estimate latent factors
of the objects. \emph{Perception-prediction network (PPN)} \cite{Zheng2018-bk}
is another attempt at estimating latent physics factors by using Graph
Neural Networks (GNNs) to build an inductive bias by observing object movement dynamics. DensePhyNet \cite{DBLP:conf/rss/Xu0ZTS19}  extends this concept by allowing iterative actions on objects to learn a latent representation using a deep learning architecture that is capable of mapping to latent physics factors of the objects. We use a similar
idea as \emph{Galileo}, but we evaluate latent factors in a task oriented
manner by combining ideas from transfer learning in RL \cite{fernandez2006probabilistic, zhu2020transfer}.

When training robots in simulated environments for transfer, domain
randomisation \cite{8202133,DBLP:conf/rss/SadeghiL17} and domain
adaptation \cite{10.5555/3306127.3331943} are generally the two
main approaches adopted. Domain randomisation generally trains robots
with randomised latent factors, whereas domain adaptation attempts
to match the data distribution in the simulator to the real-world
through techniques such as regularisation. 

Grounded learning\cite{farchy2013humanoid}
is a domain adaptation technique that attempts to train robots in
a simulated environment by attempting to reduce the trajectory difference
between simulated and the real worlds. Various extensions of this concept have been used to identify system parameters in robotic applications \cite{TossingBot, allevato2020tunenet}. In this study, while adopting
a simulation based approach similar to grounded learning, we focus
on the effects of using knowledge of physics to select better actions
that expedite the process of physics latent factor estimation.

Intrinsically motivated exploration methods such as curiosity have
proven to be effective, especially when the reward is sparse \cite{10.5555/3305890.3305968}.
Denil et al. \cite{DBLP:conf/iclr/DenilAKEBF17} have developed an
RL agent to answer physics related questions by learning to explore
without any physics prior being given. Zhou et al. \cite{zhou2018environment}
have used deep RL methods to select object trajectories that are most
useful in predicting target trajectories. Consistent with the overall
goal of these studies, our method also aims to improve sample efficiency
through improved exploration, albeit through the preferential selecting
action types based on physics knowledge.

\section{Approach and Framework \label{sec:framework}}
In this work, similar to the setup used in PHYRE \cite{Bakhtin2019-dq}, we consider physics based tasks where a single action is executed to solve the task under consideration. Although this corresponds to a contextual bandit setting \cite{li2010contextual}, we train the agent using RL, as it has been shown to be effective for such problems \cite{Bakhtin2019-dq}. Accordingly, we formalize our problem as a Markov Decision Process (MDP) $\mathcal{M}$, represented as a tuple $<\mathcal{S},\mathcal{A},\mathcal{R},\mathcal{P}>$, where $\mathcal{S}$ is the state space, $\mathcal{A}$ is the action space, $\mathcal{P}$ is the transition probability function and $\mathcal{R}$ is the reward function. 
In this work, we consider physics based tasks in which an RL agent receives an initial frame of the scene (i.e. state) $s\in \mathcal{S}$, executes an action $a\in \mathcal{A}$, which results in a reward $r\in \mathcal{R}$. The objective is for the agent to interact with the environment to learn a policy that maximises the expected rewards.

In physics based tasks, the transition function $\mathcal{P}$ depends on: (a) modelable latent factors $L$ (mass, coefficient of restitution, friction, etc.) associated with the various objects involved (b) unpredictable factors $U$ specific to the scenario (eg: wind resistance) and (c) the underlying rules $\rho$ that govern the dynamics of the physical world (eg: Newton's laws of motion). Thus, the transition function can be represented as: 
\begin{equation}
\mathcal{P}=\rho(\phi)
\end{equation} 
where $\phi=[L,U]$.
The benefit of the above representation is that $\rho$ is universally consistent, and can thus be incorporated as a prior. For instance, in our approach, we encode this prior by means of a physics simulator, which inherently accounts for our knowledge of object interactions. The incorporation of such prior knowledge allows one to efficiently build a model of the real world in a structured manner, by leveraging the universal nature of $\rho$.

Similar to the task in the real world, a simulated task can also be framed as an MDP $\mathcal{M}_{sim}=<\mathcal{S},\mathcal{A},\mathcal{P}_{sim},\mathcal{R}>$. Such an MDP shares the same state space $\mathcal{S}$, action space $\mathcal{A}$ and reward function $\mathcal{R}$ as the real world MDP $\mathcal{M}$, and differs solely in the transition function. 

The transition function used in simulation is given by $\mathcal{P}_{sim}=\rho(\theta)$, where $\theta$ represents latent factor estimates corresponding to an ideal simulation environment devoid of the unpredictable factors such as wind resistance. The intuition is that unpredictable factors, due to its nature, may vary drastically in different scenarios. Hence, by learning the equivalent latent factors $\theta$ in an ideal simulated environment, one can proceed to approximate $\mathcal{P}$ without any inherent assumptions about the environment in question. Differences between the unpredictable factors in the real and simulated environments may be at least partially compensated for through the learned values of $\theta$. In the following section, we describe in detail our proposed approach to determine $\theta$:

\subsection{\label{subsec:latent_est}Latent Factor Estimation}

In IPLW, first the real world trajectories $\tau$ are generated using action selection strategies, described in Section \ref{subsec:Action-Selection-Strategies}.
In order to obtain reasonable estimates of latent factors that minimize the difference between $\mathcal{P}$ and $\mathcal{P}_{sim}$, we then generate simulated trajectories $\tau_{sim}$ corresponding to some sampled latent factors $\theta$. By comparing $\tau_{sim}$ to real-world trajectory $\tau$, we can compute the residual as:  \begin{equation} \nabla(\theta)=\frac{1}{|D|}\sum_{\tau\in D,\tau_{sim}}d(\tau,\tau_{sim}) \end{equation} 
Here, $d$ is measured as the sum of squared difference between the states in the simulated and real-world trajectories. 

The computed residual is then minimized using an optimiser to yield updated values for the latent factors:
 \begin{equation}
   \label{eqn:theta_estimate}
  \theta\leftarrow\underset{\theta}{argmin}\nabla(\theta) 
  \end{equation} 
        In this study, the optimiser used is the cross-entropy method \cite{de2005tutorial}. By iteratively updating $\theta$ as per the above equation, we obtain improved estimates of $\theta$ that allows $\tau_{sim}$ to closely match the sampled trajectories $\tau$.
        
\subsection{\label{subsec:Action-Selection-Strategies}Action Selection Strategies}
In rolling out trajectories to estimate $\theta$, the type of actions chosen could be an important choice, depending on the scenario. For example, for a task that is primarily dependent on rolling actions, rolling out collision-heavy trajectories without much rolling is not likely to be useful. This is because collision-heavy trajectories are unlikely to improve friction-related latent estimates. Similarly, tasks that rely heavily on CoR estimates are unlikely to benefit from action types that do not produce collisions. Hence, the choice of action types can significantly influence latent factor estimates. 
Using this intuition, we design a number of action selection strategies (Algorithm \ref{alg:iplw}, line 3) that could be used for latent factor estimation:
\begin{enumerate}
\item Collisions - actions that cause the most number of collisions among balls are preferentially selected. 
\item Rolling - the total number of timesteps each ball keeps in contact with the floor during a simulation is counted. Under this strategy, actions with the highest rolling times are preferentially selected. 
\item Random - actions are selected randomly without any consideration to rolling times or collisions.
\item Mixed - collision and rolling actions are selected in equal proportion.
\item Gradient based - Ideally, action types should be selected in accordance to the performance improvements they facilitate. If an estimate of the transfer performance ($\mathcal{J(\theta)}$) as a function of the latent factors $\theta$ is available, it can be used to generate
selection probabilities (eg: probability of choosing rolling actions) corresponding to each action type. For example, if $\theta=[\theta_{CoR}, \theta_{friction}]$, the probabilities of selecting collision and rolling action types can be given by $\frac{\partial \mathcal{J}(\theta)}{\partial \theta_{CoR}}$/($\frac{\partial \mathcal{J}(\theta)}{\partial \theta_{CoR}}$+$\frac{\partial \mathcal{J}(\theta)}{\partial \theta_{friction}}$) and $\frac{\partial \mathcal{J}(\theta)}{\partial \theta_{friction}}$/($\frac{\partial \mathcal{J}(\theta)}{\partial \theta_{CoR}}$+$\frac{\partial \mathcal{J}(\theta)}{\partial \theta_{friction}}$) respectively. 
This strategy samples action types in accordance to the
relative importance of different latent factors for achieving a task. The gradient based strategy is summarised in Algorithm \ref{alg:gradients}. 
\end{enumerate}

\begin{algorithm}[t]
\caption{Gradient Based Action Selection}
\label{alg:gradients}
\hspace*{\algorithmicindent} \textbf{Input:}
\begin{algorithmic}[1]
\State $\hat{\theta}$: $\mathbb{R}^N$ vector of estimated latent factors
\State $\mathcal{J}(\theta):$ function that returns jump start performances for latent factors $\theta$
\State \textbf{Output:} $P:$ Vector of action type selection probabilities corresponding to $N$ latent factors
\State $X \gets []$
\For{$k \gets 1$ to $N$}                  
$X_k \gets \frac{\partial \mathcal{J}(\theta)}{\partial \theta_k}$ evaluated at $\hat{\theta}$
\EndFor
\State $Z \gets \sum_{n=1}^{N} X_n$

\State return $P=[\frac{X_1}{Z}, ... ,\frac{X_N}{Z}]$ 
\end{algorithmic}
\end{algorithm}

\subsection{\label{subsec:Transfer_mech}Transfer Mechanism:}

Using the latent factor estimate $\theta$ in Equation \ref{eqn:theta_estimate}, we can approximate $\mathcal{P}_{sim}$ as closely as possible to $\mathcal{P}$. The corresponding model is assumed to be closely related to the the real world, and it thus facilitates the learning of a good policy, solely using the simulator. This enables the offloading of a bulk of the learning to the simulated world, thereby drastically reducing the number of real world samples required for learning.  Specifically, we learn a policy $\pi^{*}$ within the simulated environment, and use the parameters of this learned policy to initialize learning in the real world. If $\pi^{*}$ corresponds to a good policy, we expect it to result in jump-start improvements in the real world learning performance. The overall algorithm (IPLW), is summarised in Algorithm \ref{alg:iplw}.

\begin{algorithm}[t]
\caption{Intuitive Physics Learning Workflow}
\label{alg:iplw}
\hspace*{\algorithmicindent} \textbf{Input:}
\begin{algorithmic}[1]
\State $E$, $E_{sim}$ - Real and simulated environments
\State $min\_residual>0$, $max\_attempts>0$ - Interrupt thresholds
\State $action\_selection\_strategy$ - Strategy to select an action type (e.g. collisions, rolling .etc)
\State $optimizer$ - The CEM optimizer used to improve latent factors
\State $N$ - Number of actions to use for $optimizer$
\State // initialize
\State $\theta \gets Random()$; $num\_rounds \gets 0$
\State \textbf{Output:} Simulation policy $\pi^{*}$

\While{True}
	\State $exploring\_actions$ $\gets$ Sample actions on $E_{sim}$ initialized with $\theta$
	\State $exploring\_actions$ $\gets$ Rank $exploring\_actions$ according to $action\_selection\_strategy$ and select best $N$ actions
	\State $D$ $\gets$ Rollout $exploring\_actions$ on environment $E$ and collect real-world trajectories
	\State $\theta_{sampled}$, $residuals$ $\gets$ Sample $\theta$ in $E_{sim}$ with $exploring\_actions$
	\State $\theta$, $residual_{min}$ $\gets$ $optimizer$$(\theta_{sampled}$, $residuals)$
	\State Increment $num\_rounds$
    \If {$residual_{min} < min\_residual$ or $num\_rounds$ > $max\_attempts$}
		\State Train DQN on $E_{sim}$ with $\theta$ to learn policies $\pi^{*}$
		\State Transfer trained policies $\pi^{*}$ to real-world $E$
		\State Break
	\EndIf
\EndWhile
\end{algorithmic}
\end{algorithm}

\section{Experiments\label{sec:Experiments}}
\subsection{Experiment Setup}

\begin{figure*}
\begin{centering}
\subfloat[Basketball - throw the ball into the basket using a plank]{\begin{centering}
\includegraphics[width=0.25\paperwidth]{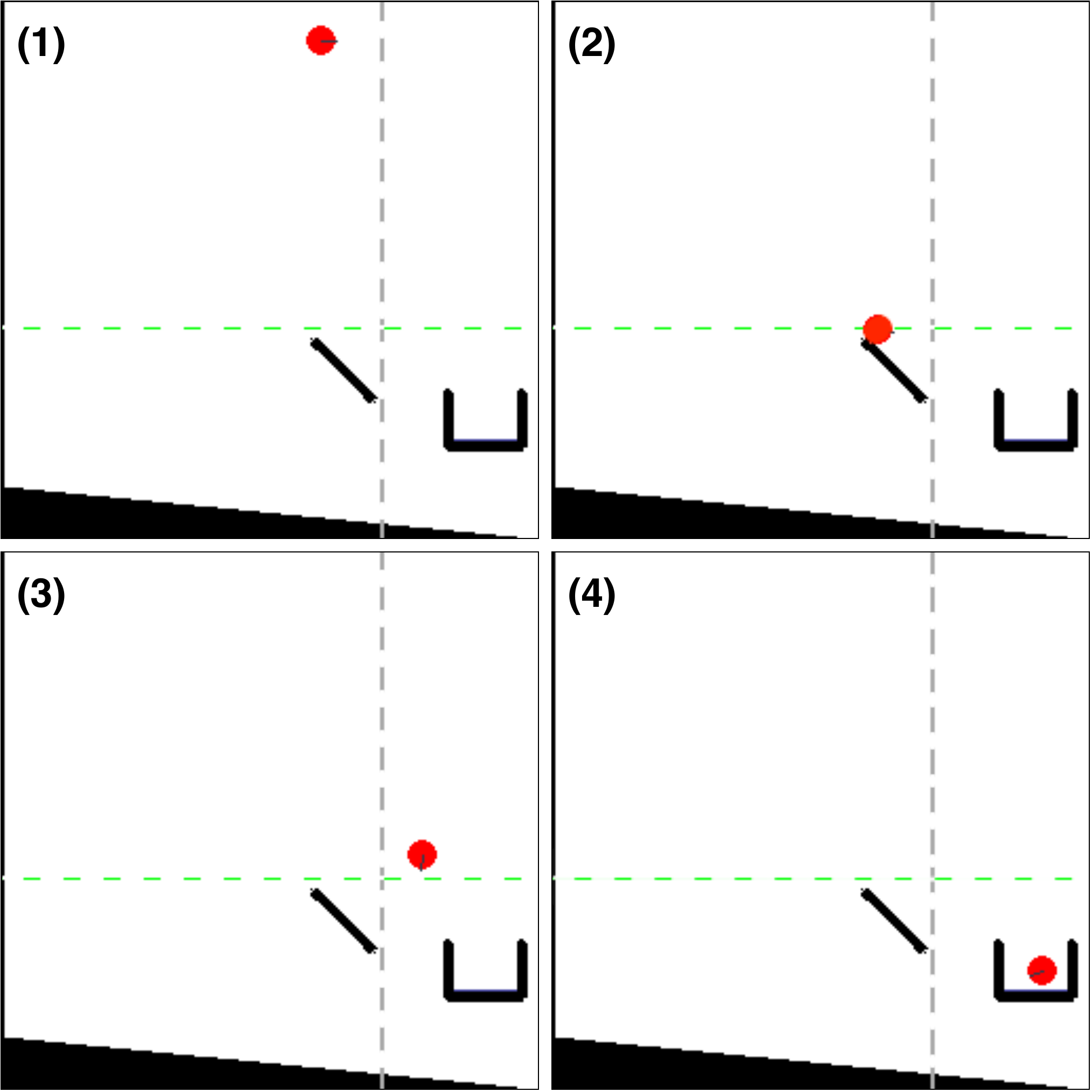}
\par\end{centering}
}\hspace{0.01\textwidth} 
\subfloat[Bowling - keep the green and blue balls in contact]{\begin{centering}
\includegraphics[width=0.25\paperwidth]{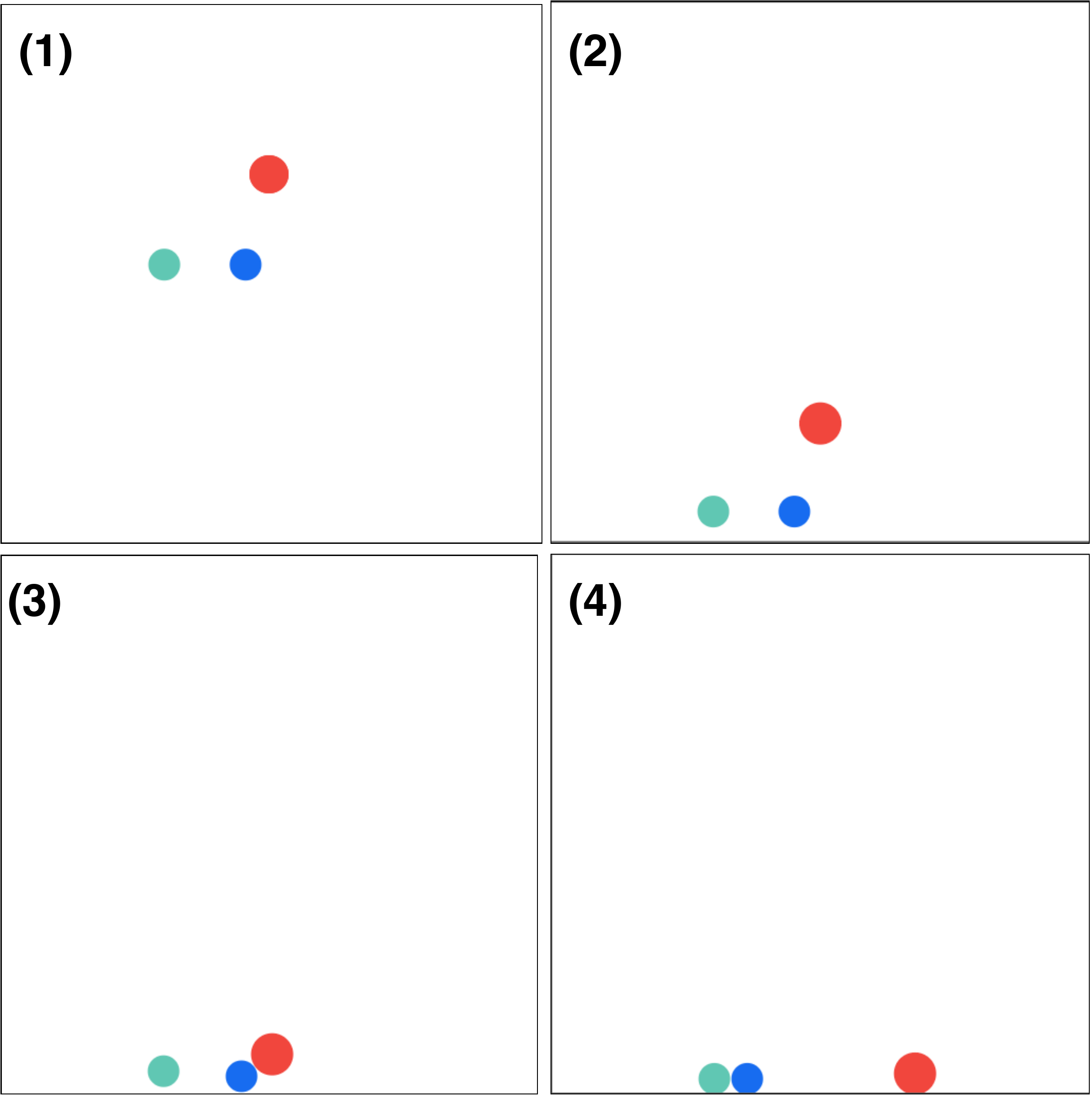}
\par\end{centering}
}
\par\end{centering}
\caption{\label{fig:basketball}Physics tasks for evaluating the sample efficiency of RL agents. 1) The ball and the plank (fixed at 45 degree angle) need to be positioned so that the ball bounces into the basket. The plank position is constrained to be below the horizontal green line and to the left of the vertical gray line. 2) Green and blue balls are initialized with different locations and sizes. An agent can choose the position ($x$, $y$ coordinates) and the radius of the red ball with the aim of keeping the blue and green balls in contact for 3 seconds.}
\end{figure*}

To demonstrate the general nature of our approach, we select two distinct physics based tasks: (a) Basketball, a collision intensive task, and; (b) Bowling, a task involving both rolling and collision actions. In the task (a), a free falling ball must be positioned together with a plank fixed at 45 degrees, so that the ball bounces off the plank and into a basket set up some distance away (Figure \ref{fig:basketball} (a)). For the task (b), we consider a 0000 PHYRE template \cite{Bakhtin2019-dq} based task consisting of 3 balls, where the goal is to keep two of the balls in contact by controlling the position and size of the red ball (i.e. action) (Figure \ref{fig:basketball} (b)). 

 For the purposes of this study, both the simulated and real-world
are generated using the Pymunk physics simulator\footnote{http://www.pymunk.org/}.
The simulated world is set up to differ from the real world through
a different value of a ``damping'' factor. The damping factor introduces
an additional drag force on objects, i.e. a value 1 represents no
drag while 0.9 means each moving body will lose 10\% of its velocity
per second. In the context of real-world forces, the damping factor
represents physics factors that are difficult to model universally,
e.g. rolling friction, air-resistance.

The overarching goal is to determine a policy trained in the simulator, that can transfer well to the real world to facilitate improvements in the real world learning performance. The performance is evaluated in terms of an \emph{AUCCESS} score, which is a weighted average score that emphasizes the speed of learning. Given a range of attempts $k\in\lbrace1,..,100\rbrace$ at solving a given task, we determine weights $w_{k}$ as $w_{k}=log(k+1)-log(k)$, and compute the \emph{AUCCESS} score as \textbf{ AUCCESS $=\sum_{k}w_{k}.s_{k}/\sum_{k}w_{k}$} , where
$s_{k}$ is the success percentage at $k$ attempts \cite[Sec. 3.2]{Bakhtin2019-dq}.

\subsection{Effect of Latent Factor Estimation\label{subsec:Domain-Randomisation-vs}}

To illustrate the benefits of latent factor estimation, we consider the Bowling task (Figure \ref{fig:basketball} (b)). Here, the damping factor is set to $0.8$ in the real environment, whereas the simulated environment assumes an ideal damping factor of $0$. Following the procedure described in Section \ref{subsec:latent_est}, with using a random action selection strategy, we first sample $10,000$ trajectories in the real world, and use a cross-entropy (CE) method optimizer to estimate latent factors. We assume a homogeneous object composition, and additionally, that the RL agent has the ability to sense relative measurements between balls (e.g. relative weights, bounce) to constrain sampling.

The estimated latent factors are then used to train an agent in the simulator, using a DQN architecture to that used in PHYRE\cite[Sec. 4.1]{Bakhtin2019-dq}. For training, we use 1000 batches of size 16 to pretrain, after which the resulting policy is transferred to the real environment to measure the jump start improvement.

We compare latent estimate based policy to the following baselines:

\begin{itemize}
    \item \emph{Domain Randomization(DR):} Here, latent factors are randomly sampled, and the policy is trained on the corresponding variety of simulation environments. The latent factors are randomized by uniformly sampling mass, friction and CoR from the ranges [0.1,20.0],[0.0,3.0] and [0.0,1.0] respectively.
        \item \emph{True Latent:} This corresponds to an oracle case, where the simulation agent is trained based on the true values of the latent factors. The true latent factors are set as: density: 0.25, friction: 0.707 and CoR: 0.447.
        \item \emph{Direct Learning:} Here, the agent learns purely through real world interactions, without the use of simulated environments.
\end{itemize}

\begin{figure*}
\begin{centering}
\hspace{0.01\textwidth} \subfloat[Transfer of pre-trained batch size 16 - transfer batch size 32]{\begin{centering}
\includegraphics[width=0.35\paperwidth]{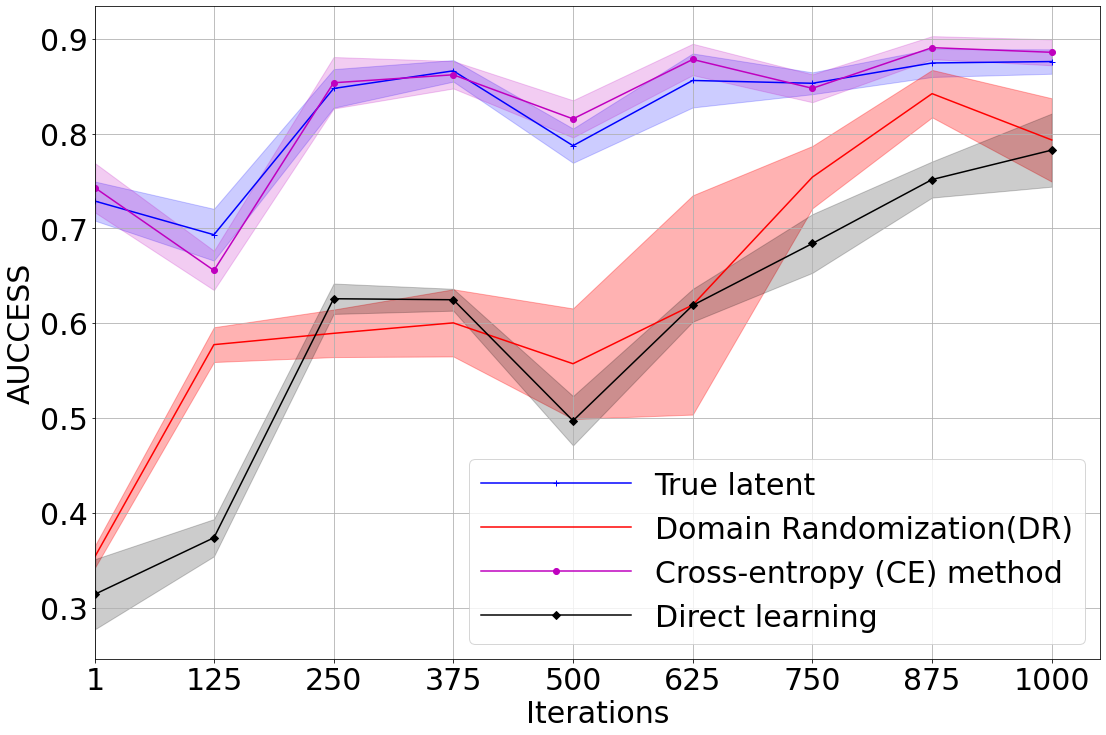}
\par\end{centering}
}
\par\end{centering}
\caption{\label{fig:Comparing-best-latent}Comparing best latent estimate vs domain randomisation based pretraining. For DR, CE and true latent, 5 pre-training trials followed by a transfer trial to the real-world are run, and AUCCESS is evaluated at each 125 batch interval and averaged. For direct learning, 5 trials are trained in the real-world and averaged. Shaded regions depict the standard error.} 
\end{figure*}

Figure \ref{fig:Comparing-best-latent} depicts the real world learning performances of the different baselines, all of which, except `Direct Learning' initializes learning with a policy learned in simulation. From the learning curves, we can infer that the knowledge of latent factors, either true or estimated, is generally more effective than DR or direct learning. Although the various approaches tend to converge to a similar final performance, the knowledge of latent factors is seen to drastically improve the jump start  \cite{lazaric2012transfer} learning performance.
These results strongly support the idea of estimating latent factors in order to learn reasonably accurate pretrained policies to initialize learning in the real environment.

\subsection{Effect of Action Type Selection}

The previous section confirmed the potential of latent factor estimation for obtaining policies that result in good jump start performances in the real environment. 
However, one major weakness of latent factor estimation in Fig. \ref{fig:Comparing-best-latent} is the large number of interactions required to arrive at satisfactory estimates. We posit that the action selection strategies listed in Section \ref{subsec:Action-Selection-Strategies} would aid in efficiently obtaining these estimates, as they leverage the defined action groupings to explore specific action types during learning.

In this section, we consider both Basketball and Bowling tasks (Figure \ref{fig:basketball}), in which, each of the action selection strategies (collisions, rolling, random, mixed and gradient based) is allowed to interact with the real environment for $500$ interactions, based on which the latent factors are estimated using IPLW. The obtained latent factor estimates are then used to learn policies in simulation, which are then transferred to the real environment to measure the jump start improvement in learning performance. 

\begin{figure*}
\begin{centering}
\subfloat[]{\begin{centering}
\includegraphics[width=0.5\paperwidth]{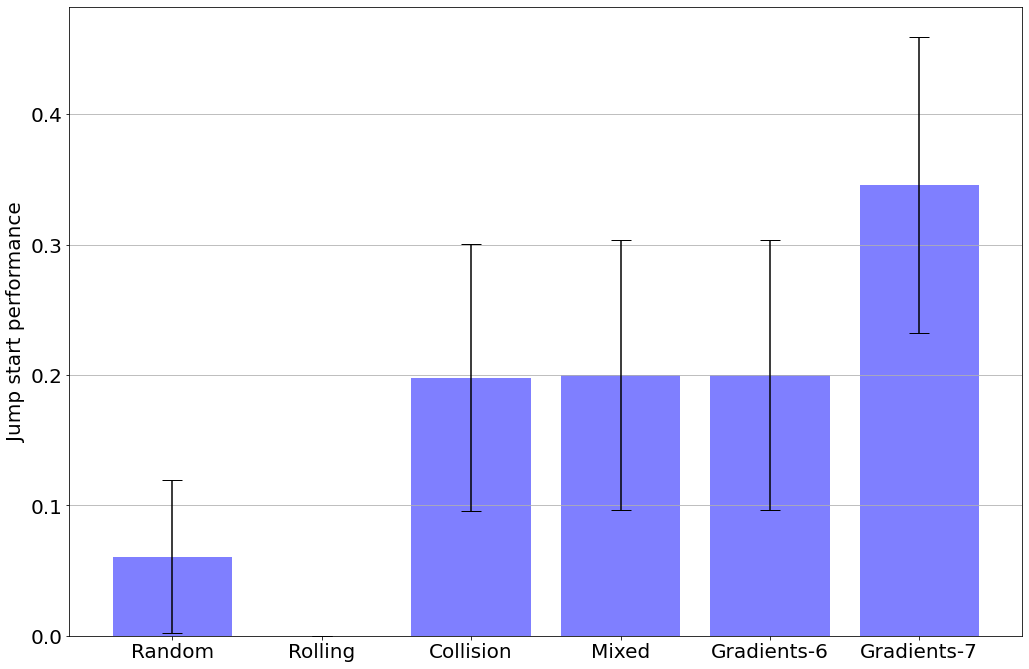}
\par\end{centering}
}\hspace{0.01\textwidth} \subfloat[]{
\begin{centering}
\includegraphics[width=0.3\paperwidth]{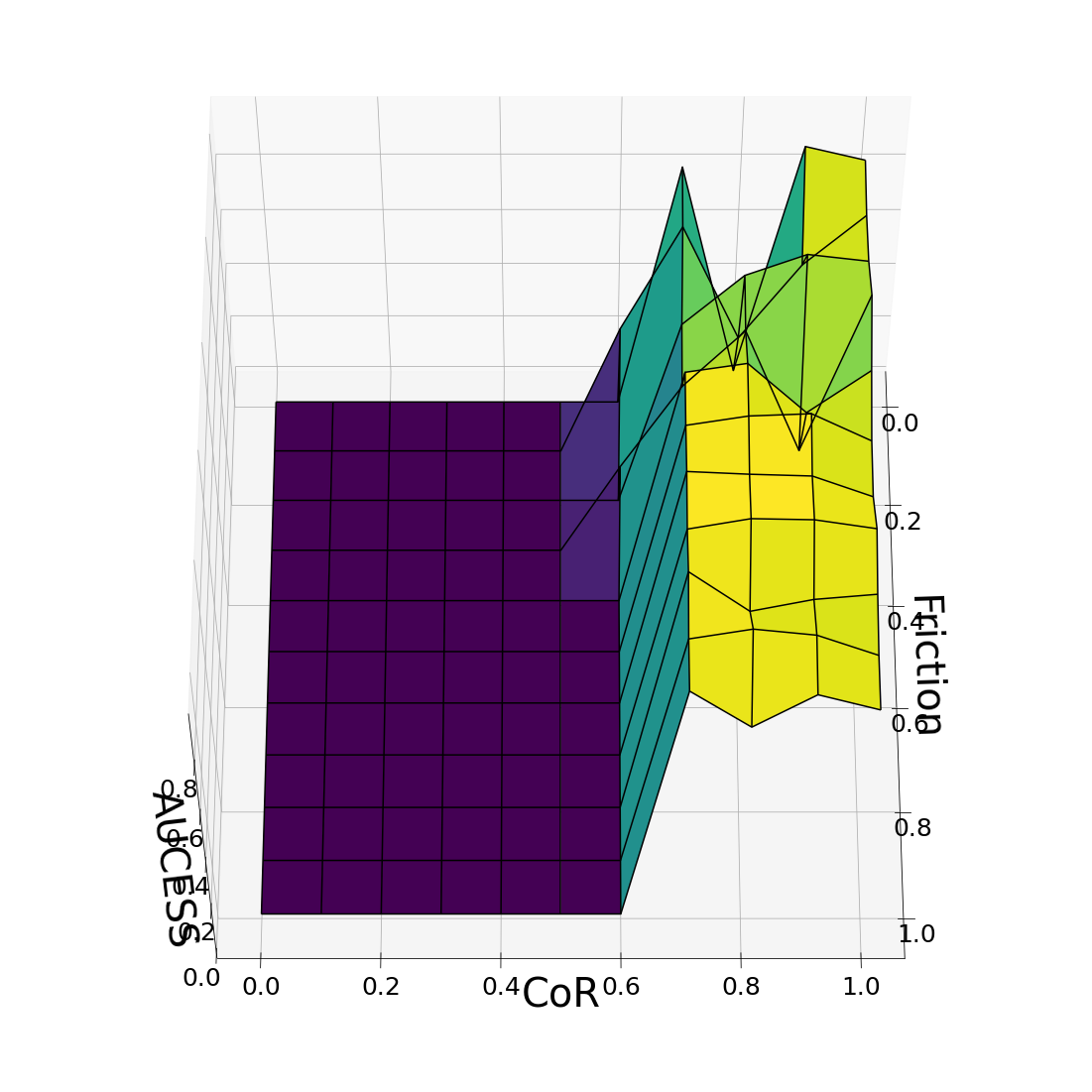}
\includegraphics[width=0.3\paperwidth]{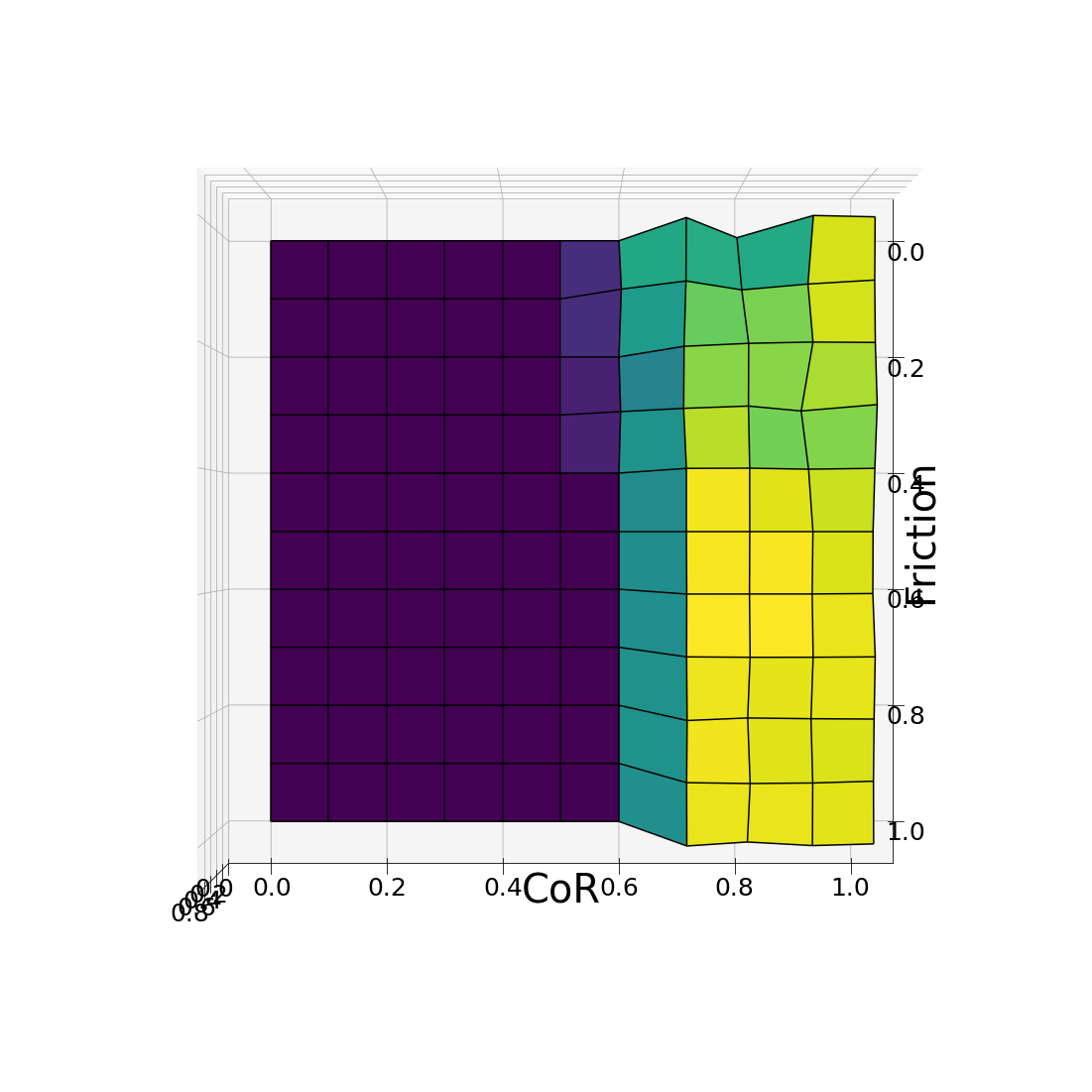}
\par\end{centering}
}
\par\end{centering}
\caption{\label{fig:Sample-efficiency-of-task1}Evaluating IPLW based learning for Basketball task. (a) Evaluating action selection strategies for transfer on validation tasks, using 10 iterations of 50 actions/iteration. Gradient-7 and Gradient-6 refer to damping=0.7 and 0.6 environments respectively. (b) Jump start performances of friction vs. CoR combinations in damping 0.7 simulated environment. Yellow to blue indicates high to low values.}
\end{figure*}

Figure \ref{fig:Sample-efficiency-of-task1} (b) shows the transfer performance surface for the Basketball task. As inferred from the figure, this task requires a significant level of accuracy in the CoR estimate, whereas the friction estimate is inconsequential (real-world latent values, friction 0.2, CoR 0.7). This is further confirmed in the jump start transfer performance plots in Figure \ref{fig:Sample-efficiency-of-task1} (a), where estimates from rolling actions achieve 0 performance, whereas those from collisions appear to significantly improve the performance. Compared to random exploration, Gradient-7 and Gradient-6 plots demonstrate the efficacy of biasing action selection using gradient ratios from relatively similar environments. 

\begin{figure*}
\begin{centering}
\subfloat[]{\begin{centering}
\includegraphics[width=0.5\paperwidth]{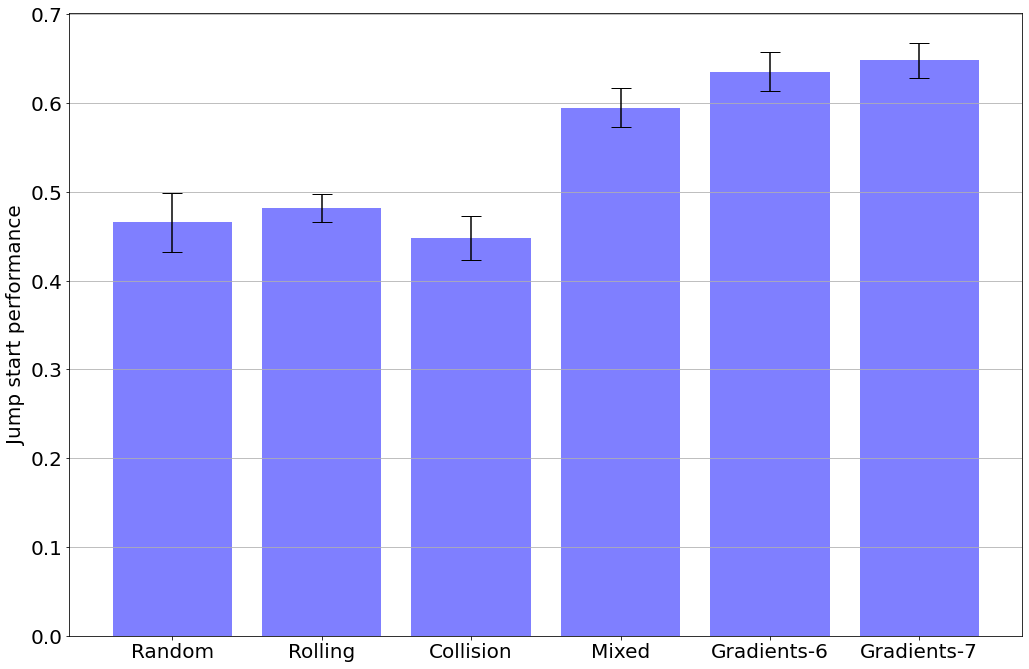}
\par\end{centering}
}\hspace{0.01\textwidth} \subfloat[]{\begin{centering}
\includegraphics[width=0.3\paperwidth]{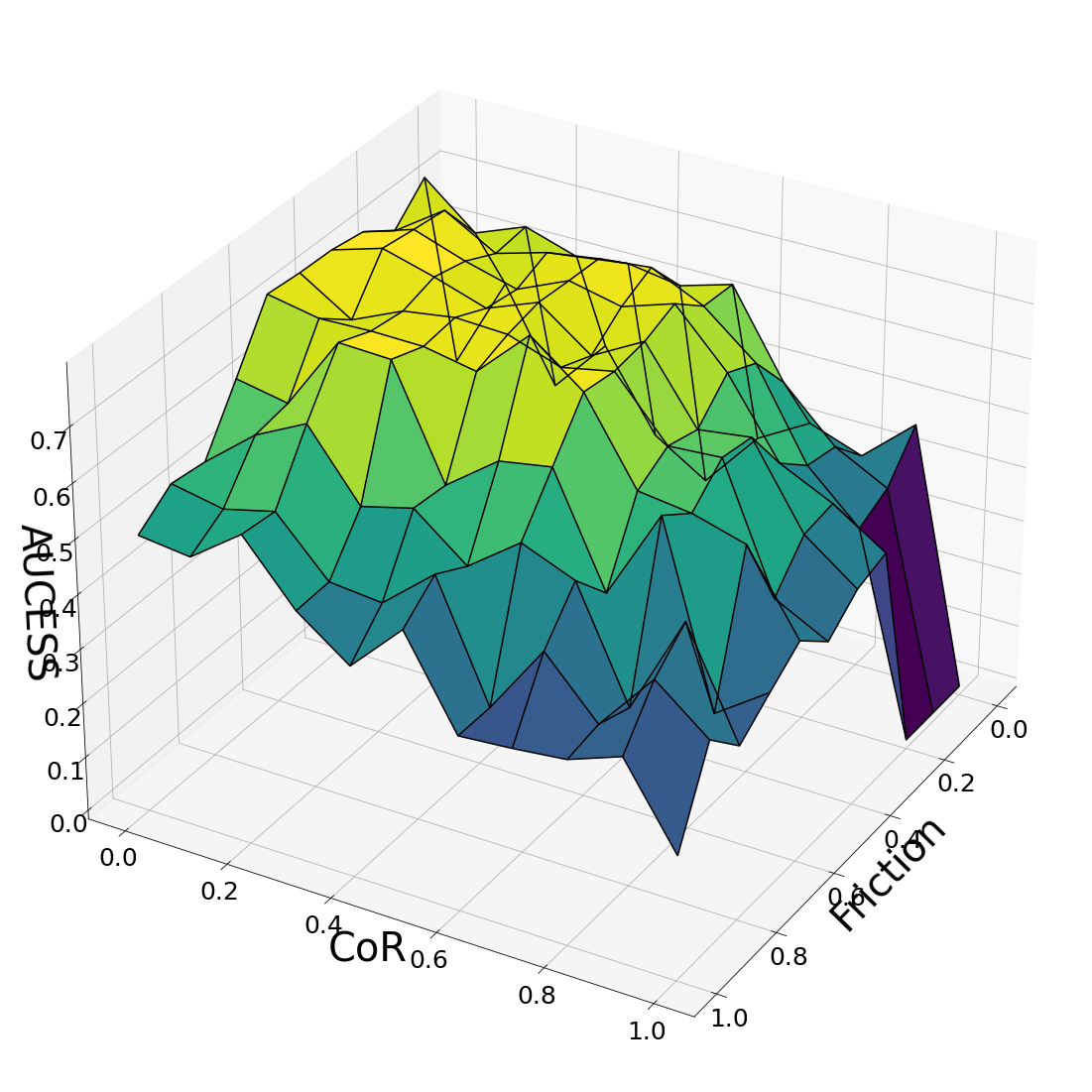}
\includegraphics[width=0.3\paperwidth]{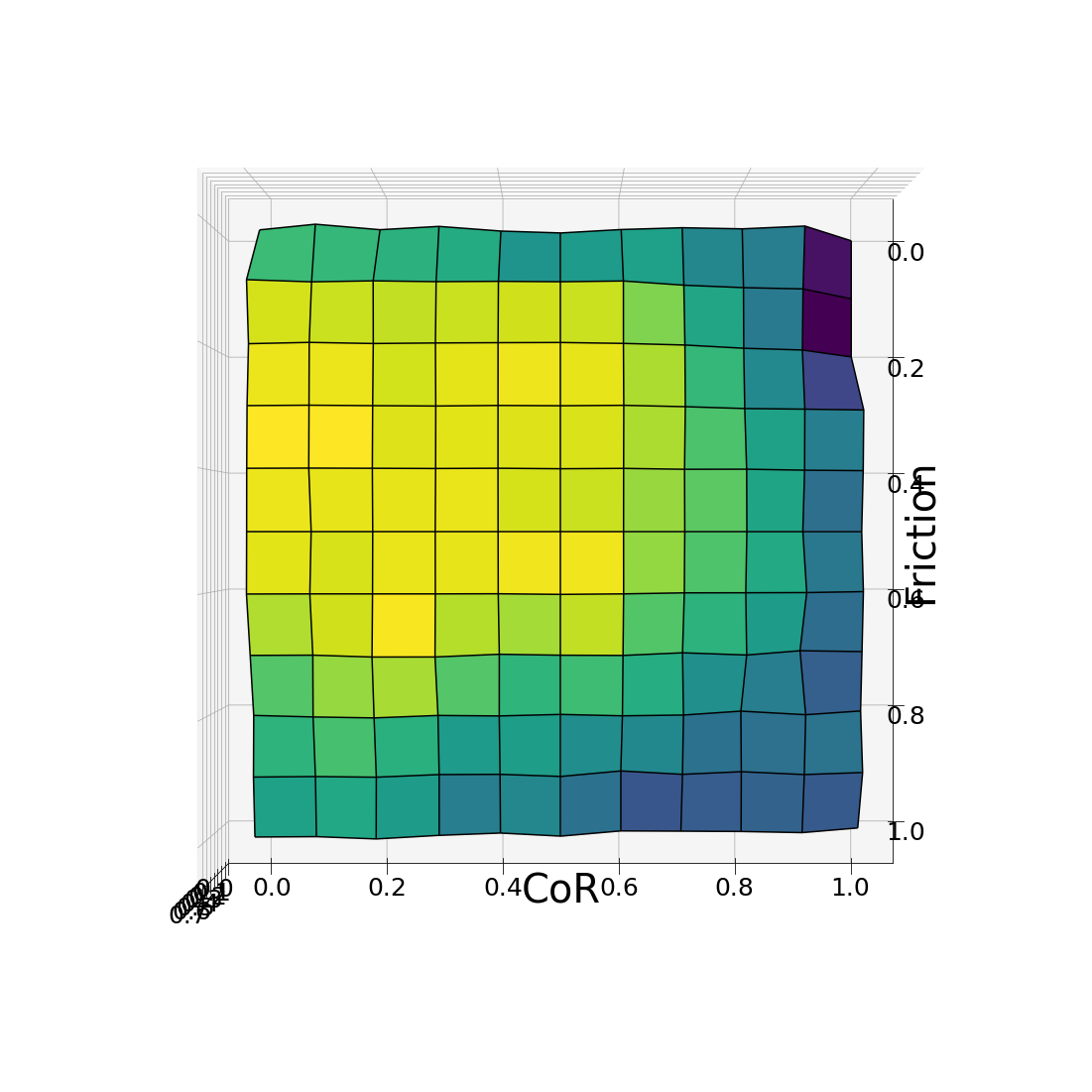}
\par\end{centering}
}
\par\end{centering}
\caption{\label{fig:Sample-efficiency-of-task2}Evaluating IPLW based learning for the Bowling task. (a) Evaluating action selection strategies for transfer on validation tasks, using 10 iterations of 50 actions/iteration. Gradient-7 and Gradient-6 refer to damping=0.7 and 0.6 environments respectively. (b) Jump start performances of friction vs. CoR combinations in damping 0.7 simulated environment. Yellow to blue indicates high to low.}
\label{fig:perf2}
\end{figure*}

The corresponding jump start plot for the Bowling task is depicted in Fig. \ref{fig:Sample-efficiency-of-task2} (a), and suggests that the gradient based methods perform the best, followed by the mixed strategy. Action selection strategies that used only one action type (rolling or collisions) for estimation performed relatively worse, probably due to the fact that these strategies led to some of the latent factors to be poorly estimated.

Among the action selection strategies, the gradient based strategy relied on previously obtained characteristics of transfer performances as a function of the different latent factors. Although such a detailed characteristic may be expensive to obtain in terms of the number of real world interactions, it may allow one to select appropriate action types irrespective of the unmodeled factors that vary across environments. This is illustrated by the `Gradient-7' performance, which despite applying the gradient based approach in an environment with damping $0.7$, was able to achieve good jump start improvements in the real environment, which is associated with a damping of $0.8$. This shows the ability of the gradient based approach to leverage trends in the transfer performance characteristics for a given task, to achieve good transfer performances in other environments, irrespective of their associated unmodeled factors.

\subsection{Implementation Details}

We use cross-entropy (CE) method as the optimizer in IPLW, using 10 rounds of 1000 multivariate normal sampled mass, friction and COR estimates with initial means and standard deviation respectively (1.0, 0.5),(1.0, 0.5) and (0.5, 0.25). At each round, we consider 200 best ranked samples (e.g. elite samples) from which the latent factor distribution is estimated, and used to sample latent factors for the next round. For all following results, we take 5 IPLW based estimates using different random number generator seeds (50, 100, 150, 500, 1000), run 3 RL trials from each estimate and average the jump start performances from all agents. In both games, the real world differs from the simulated environment through the \emph{damping} factor, which is set to $0.8$ for the real world and 0 for the simulated. During latent value estimation, 10 rounds of iterations are used ($n_{max}=10$) in IPLW.


\emph{Basketball Task:}
Contains 25 total tasks, with varying basket positions, split into 15 training, 5 validation and 5 test tasks. For the action space, we use 40,000 discretised actions, where the ball and the plank each can take 200 positions. If an action does not achieve the goal, a trajectory produces 10 observations corresponding to 500 time steps observed at 50 step intervals, whereas completion of the goal will terminate the trajectory immediately. For the purpose of this study, we use a simplified scenario where both the ball and the plank are made of the same material, with friction=$0.2$ and CoR=$0.7$, and the agent is aware of the latent factor values of rest of the objects in the task (e.g. walls, floor, basket).  In addition, we use an initial normal distribution with mean and standard deviation set to (0.5, 0.25) for the friction estimate in the CE optimiser. When pretraining the DQN, we use 1000 of 32 size batches, for both IPLW and gradient plots (Fig. \ref{fig:Sample-efficiency-of-task1} (b)).

\emph{Bowling Task:}
Contains 100 total tasks, with varying green and blue ball positions and sizes, split into 60 training, 20 validation and 20 test tasks. For this task we consider an action space of 50,000 discretised actions. Each trajectory contains 68 observations corresponding to 4000 time steps observed at 60 step intervals, equivalent to approximately 16.6 seconds of real-world play in total length. When pretraining the DQN, we use 1000 of 16 size batches, for both IPLW and gradient plots (Fig. \ref{fig:Sample-efficiency-of-task2} (b)).

\section{Discussion}
From a broad perspective, in this study, we try to decompose the complex relationship between the action types, the latent factor information associated with an action type and the latent factor information most relevant to a given task. A standard approach is to iteratively learn a simulated policy, deploy it in the real-world and minimise the difference between the real and simulated trajectories \cite{du2021auto, stocgat20, gat2017}. Instead, by decomposing this complexity, we believe it is possible to make solutions more generally applicable. For example, the relationship between action types and the latent information they provide can be considered as an invariant prior that can be reused for a different problem. With such a prior, solving a given task can be simplified to learning task-specific latent factor dependencies. Moreover, this type of decomposition may facilitate transferring latent factor dependencies across different task types. For example, latent factor dependencies from a sport such as Golf may be transferable to billiard sports (e.g. snooker, carom),  due to the similarity between tasks' friction dependency (rolling motion) and CoR dependency (initial strike on the ball). Such a task-specific latent factor dependency prior can be combined with an action type prior from the billiard sports domain where it uses a billiard cue to hit the ball instead of a golf club. Extending the concept, transferring from multiple such sources is an interesting open question.

When learning task-specific latent factor dependencies, our current implementation uses a considerable number of transfer performance samples to generate the performance surfaces shown in Fig. \ref{fig:Sample-efficiency-of-task1}(b),(d). However, it may not be possible to recreate similar surfaces when a large number of latent factors are involved. Such scenarios may demand uncertainty-based sampling techniques to focus on specific regions of the performance surface or function approximations. Dedicated attention mechanisms \cite{attention_all} to identify relevant objects to a task may also be useful in this regard. Interestingly, mixing action types in equal proportions can be a reasonably sample efficient and low cost exploration method. 

Apart from the latent factors, the dimensionality of the unmodelled factors (damping) may also affect Sim2real transfer. Our evaluation environments consider only a single unmodelled factor. However, in practice, these factors could be multidimensional (e.g. wind speed and direction). Although we believe a sufficient degree of similarity between real and simulated environments would ensure the handling of an arbitrary number of unmodelled factor dimensions, it presents an open issue and a potential topic for future work.

\section{Conclusion}
In this work, we proposed an approach to efficiently explore physics-based tasks based on task-specific estimates of latent factors. We introduced the ideas of intuitive action groupings and partial grounding, which formed the basis for IPLW, our proposed workflow to achieve improved exploration during learning. We empirically evaluated our proposed approach in two types of physics based learning environments and demonstrated its superior sample efficiency compared to various baselines. This type of structured exploration could be combined with curiosity based exploration to develop agents capable of adapting to new environments using only a handful of interactions.

\pagebreak

\bibliographystyle{splncs04}
\bibliography{physics}

\end{document}